\documentclass[10pt,twocolumn,letterpaper]{article}

\usepackage{cvpr}              

\definecolor{cvprblue}{rgb}{0.21,0.49,0.74}
\usepackage[pagebackref,breaklinks,colorlinks,allcolors=cvprblue]{hyperref}
\usepackage{multirow}
\usepackage{array}
\def\paperID{1173} 
\def\confName{CVPR}
\def\confYear{2025}

\title{Collaborative Decoding Makes Visual Auto-Regressive Modeling Efficient}

\author{
    Zigeng Chen
    ,~ Xinyin Ma
    ,~ Gongfan Fang
    ,~ Xinchao Wang\thanks{Corresponding author} \\
    National University of Singapore \\
    {\small \tt{\{zigeng99, maxinyin, gongfan\}@u.nus.edu, xinchao@nus.edu.sg}} \\
    \url{https://github.com/czg1225/CoDe}
}

\begin{document}

\maketitle

\begin{abstract}

In the rapidly advancing field of image generation, Visual Auto-Regressive (VAR) modeling has garnered considerable attention for its innovative next-scale prediction approach. This paradigm offers substantial improvements in efficiency, scalability, and zero-shot generalization. Yet, the inherently coarse-to-fine nature of VAR introduces a prolonged token sequence, leading to prohibitive memory consumption and computational redundancies. To address these bottlenecks, we propose Collaborative Decoding (CoDe), a novel efficient decoding strategy tailored for the VAR framework. CoDe capitalizes on two critical observations: the substantially reduced parameter demands at larger scales and the exclusive generation patterns across different scales. Based on these insights, we partition the multi-scale inference process into a seamless collaboration between a large model and a small model. The large model serves as the 'drafter', specializing in generating low-frequency content at smaller scales, while the smaller model serves as the 'refiner', solely focusing on predicting high-frequency details at larger scales. 
This collaboration yields remarkable efficiency with minimal impact on quality: CoDe achieves a \textbf{1.7x} speedup, slashes memory usage by around \textbf{50\%}, and preserves image quality with only a negligible FID increase from \textbf{1.95} to \textbf{1.98}. When drafting steps are further decreased, CoDe can achieve an impressive \textbf{2.9x} acceleration ratio, reaching \textbf{41 images/s} at 256x256 resolution on a single NVIDIA 4090 GPU, while preserving a commendable FID of \textbf{2.27}.


\end{abstract}    
\begin{figure}[t]
\centering
\includegraphics[width=3.1in]{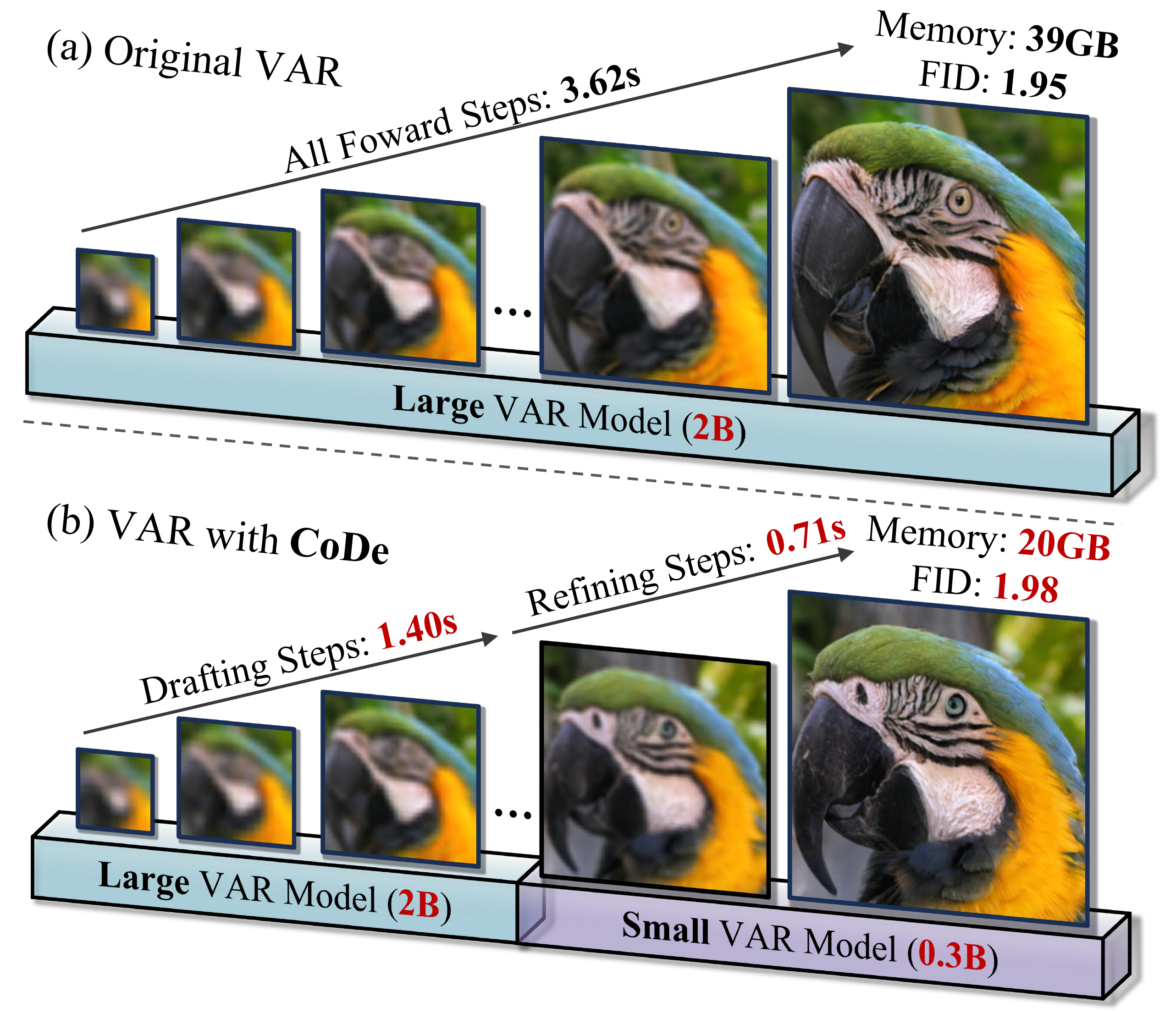}
\caption{We partition the next-scale prediction process into the efficient collaboration between large and small VAR models.}
\label{fig_intro}
\end{figure}

\section{Introduction}
\label{sec:intro}

\begin{figure*}[t]
\centering
\includegraphics[width=7.1in]{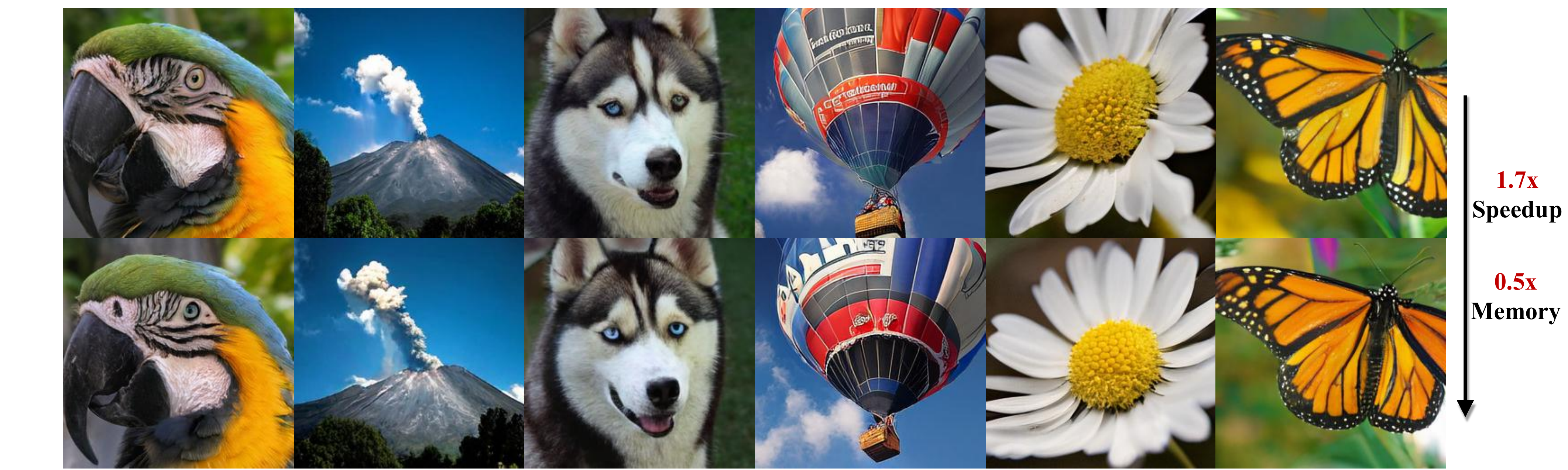}
\caption{Comparison of generation results between original VAR-d30 (up) and our VAR-CoDE (bottom) for ImageNet 256$\times$256. Our method achieves 1.7x speedup (3.62s to 2.11s), and needs only 0.5x memory space (40GB to 20GB), with negligible quality degradation.}
\label{fig_teaser}
\end{figure*}


The past year has witnessed significant advancements in \emph{Auto-Regressive} (AR) models for image generation \cite{sun2024autoregressive,li2024autoregressive,li2024scalable,yao2024car,li2024controlar,he2024mars}, driven by their proven scalability and strong generalization capabilities \cite{dubey2024llama,touvron2023llama,bai2023qwen,guo2024deepseek}. Leveraging these strengths, AR approaches have demonstrated remarkable image quality and promising multi-modal potential \cite{liu2024lumina,xie2024show,zhou2024transfusion}. However, the next-token prediction process inherent to conventional AR models requires numerous decoding steps, leading to considerable generation latency.
\emph{Visual Auto-Regressive} (VAR) Modeling \cite{tian2024visual} replaces the GPT-style next-token prediction with a next-scale prediction strategy. It generates content in a multi-level, coarse-to-fine progression, enabling the model to decode multiple tokens in parallel \cite{zhang2024var,li2024controlvar,zhang2024g3pt,qiu2024efficient,tang2024hart}, thus considerably reducing the inference steps.

Despite less decoding steps, VAR's progressive scaling approach significantly increases the overall sequence length. To generate a 16x16 token image, VAR hierarchically decodes up to 680 tokens across 10 scales—2.7 times the sequence length required by conventional AR models.  As the VAR model must store the KV cache accumulated from all previous tokens, this prolonged sequence results in a substantial memory overhead, especially in the final scales. During inference, the KV cache consumes approximately 12 times more memory than the forward computation itself. For instance, generating images using VAR-d30 with a batch size of 128 demands 70 GB of memory, with 57 GB dedicated solely to KV caching, becoming the primary bottleneck. Moreover, the extended sequence length exacerbates the computational cost of self-attention, given the quadratic growth in attention map calculations.

To address these inefficiencies, we begin by analyzing the specific properties of VAR's next-scale prediction paradigm. First, we observe that as generated scales increase, the parameter demands for high-quality token generation drop substantially, leading to considerable computational redundancy at most of the tokens in the long sequence. Next, we find that the generation patterns at small and large scales are exclusive, resulting in mutual interference between learning low- and high-frequency modeling capability, thus hindering efficient parameter utilization.


\textbf{Our Approach.} Inspired by the above observations, we propose \emph{Collaborative Decoding} (CoDe), a simple yet highly effective method that significantly boosts the inference efficiency of VAR models while preserving generation quality comparable to the original. Our core idea is to decompose the long-sequence scaling-up process into a collaboration between two VAR models with different sizes and specialized roles to enhance efficiency. Figure \ref{fig_intro} presents the overview of our method. We use a large VAR model as a drafter for the initial small scales,  where model capacity demands are high but computations and KV cache are sparse. Conversely, we use a small VAR model as a refiner at the remaining large scales, where computations and KV cache are intensive but fewer parameters are needed. Next, we apply specialized fine-tuning to both models to address the optimization interference between scales encountered during the pre-training stage. Each model is fine-tuned exclusively on the specific scales it handles, leading to a notable performance boost with limited additional cost.

Extensive experiments validate the effectiveness of our method. As shown in Figure \ref{fig_teaser}, compared to the original VAR-d30 model, CoDe achieves a \textbf{1.7x} speedup with merely \textbf{0.5x} GPU memory consumption and negligible quality loss (FID \cite{heusel2017gans} slightly increases from 1.95 to 1.98). Notably, our approach can reach up to a \textbf{2.9x} speedup, generating \textbf{41} images (256x256) per second on a single NVIDIA 4090 GPU, while still maintaining an FID of 2.27.

In conclusion, we introduce CoDe, a novel decoding framework for visual auto-regressive modeling that significantly enhances speed and reduces memory overhead with a negligible impact on quality. CoDe divides the hierarchical sequence modeling of VAR into a collaborative process, utilizing a large VAR model and a small VAR model in a progressive partnership. Additionally, we propose specialized fine-tuning to optimize each model for its specific role, effectively mitigating training interference and maximizing parameter utilization. To the best of our knowledge, CoDe is the fastest method available to achieve an FID below 2, making it a clear advancement in efficient image generation.


\section{Related Works}
\label{sec:Related Works}

\begin{figure*}[ht!]
\centering
\includegraphics[width=\linewidth]{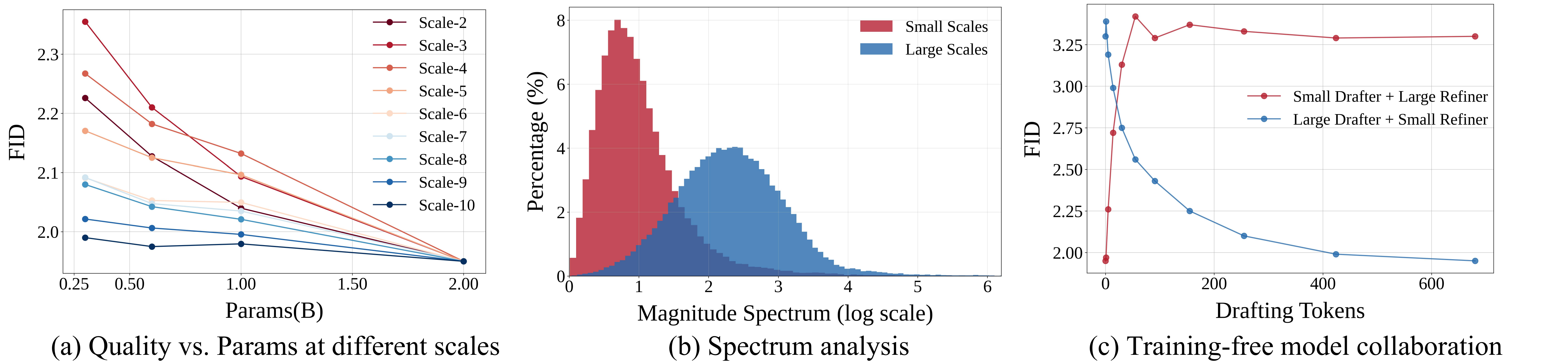}
\caption{(a) Effectiveness of increasing parameters at the $k$-th scale is evaluated by predicting token map $r_k$ using four VAR models with different parameter sizes (2B, 1B, 0.6B, and 0.3B), while other scales $(r_1, r_2, \dots, r_{k-1}, r_{k+1}, \dots, r_{10})$ are generated using the largest VAR-d30 model. (b) Fourier spectrum analysis is conducted on generated content at the first 3 scales and the last 3 scales. (c) Training-free performance comparison of model collaboration decoding across various settings of draft tokens $M$ and refiner tokens $680-M$.}
\label{fig_obs}
\end{figure*}

\noindent \textbf{Auto-regressive image generation.}
Early works \cite{chen2020generative,van2016conditional} pioneered image generation by generating pixels in raster-scan order. Later, VQVAE \cite{van2017neural} and VQGAN \cite{esser2021taming} improved this approach by quantizing image patches into discrete tokens, using a transformer in a decoder-only setup to generate these tokens in a raster-scan manner. Building on these foundations, recent efforts have focused on enhancing autoregressive (AR) models for image generation. LlamaGEN \cite{sun2024autoregressive} and Lumina-mGPT \cite{liu2024lumina}, for example, use a GPT-style next-token prediction strategy to achieve high-quality image generation with good scalability. AiM \cite{li2024scalable} and MARS \cite{he2024mars} further improve this paradigm by introducing mixture-of-experts and linear attention mechanisms \cite{gu2023mamba}. Methods like SHOW-O \cite{xie2024show}, Transfusion \cite{zhou2024transfusion}, and DART \cite{gu2024dart} combine diffusion processes with autoregressive modeling, while \cite{li2024autoregressive,chang2022maskgit,luo2024open} introduces masked autoregression to generate images. However, all these approaches suffer from high latency due to the large number of forward steps. VAR \cite{tian2024visual} addresses this by using hierarchical parallel decoding, which greatly reduces the number of steps, leading to significantly lower inference latency without sacrificing quality. Building on VAR's progress, many recent works have aimed to improve next-scale prediction across multiple tasks, including text-to-image \cite{li2024imagefolder,zhang2024var,tang2024hart,ma2024star}, controllable generation \cite{li2024controlvar,li2024controlar}, audio generation \cite{qiu2024efficient}, and 3D generation \cite{zhang2024g3pt}.

\noindent \textbf{Efficient Image Generation Models.}
For diffusion models, acceleration techniques are already well-developed.  \cite{salimans2022progressive, yue2024resshift, luo2023latent, sauer2023adversarial, yin2023one} focuses on reducing sampling steps while \cite{li2024snapfusion, zhao2023mobilediffusion, fang2024structural, zhang2024laptop, yang2023diffusion} optimize model architectures through pruning \cite{fang2023depgraph, chen20230} or knowledge distillation \cite{hinton2015distilling}. 
To avoid the high costs of training, some training-free methods are proposed. Some approaches develop fast solvers for stochastic differential equations (SDE) or ordinary differential equations (ODE) to enhance sampling efficiency \cite{lu2022dpm, bao2022analytic, liu2022pseudo, zhang2022fast, zheng2024dpm}. Other works \cite{ma2023deepcache, wimbauer2023cache, zhang2024cross, yang2024hash3d, so2023frdiff, li2023faster, lyu2022accelerating, zhao2024real} exploit specific characteristics of diffusion models to skip redundant computations during the denoising process. \cite{li2024distrifusion, chen2024asyncdiff, fang2024xdit} reduce latency to another level through distributed computing. Quantization methods \cite{li2023q,shang2023post,li2024svdquant} have also shown great potential. In contrast, research into accelerating AR image generation is still in its early stages. LANTERN \cite{jang2024lantern} and SJD \cite{teng2024accelerating} employ speculative decoding to speed up next-token prediction, achieving notable speedups but not suitable for the innovative next-scale prediction paradigm introduced by VAR.

In this paper, we introduce CoDe, a novel and efficient decoding method tailored for the next-scale prediction paradigm. CoDe effectively addresses the inefficiencies associated with the long sequence structure of VAR models while maintaining high generation quality. Notably, CoDe stands out as the fastest method capable of achieving an FID below 2, making it a clear advancement.


\section{Method}

\subsection{Prelinimary}

Visual auto-regressive modeling \cite{tian2024visual} redefines the traditional AR by shifting from a “next-token” prediction to a “next-scale” prediction. In this framework, each auto-regressive unit is a token map at varying scales, rather than a single token. For a given image feature map \( f \in \mathbb{R}^{h \times w \times C} \), VAR quantizes it into \( K \) multi-scale token maps \( R = (r_1, r_2, \ldots, r_K) \) at progressively finer resolutions, with the final token map \( r_K \) matching the original feature map's resolution. The probability distribution is reformulated as:
\begin{equation}
p(r_1, r_2, \ldots, r_K) = \prod_{k=1}^K p(r_k \mid r_1, r_2, \ldots, r_{k-1}),
\end{equation}
where each token map \( r_k \in [V]^{h_k \times w_k} \) consists of \( h_k \times w_k \) tokens at scale \( k \), and the sequence \( (r_1, r_2, \ldots, r_{k-1}) \) serves as the "prefix" for \( r_k \). In this paradigm, during each autoregressive step \( k \), the model predicts all \( h_k \times w_k \) tokens in \( r_k \) in parallel, conditioned on prior scales and position embeddings. This approach aligns with a coarse-to-fine generation pattern, enabling parallel decoding within each scale. VAR effectively improves inference speed and generation quality, but it also considerably increases sequence length.

\begin{figure*}[ht!]
\centering
\includegraphics[width=\linewidth]{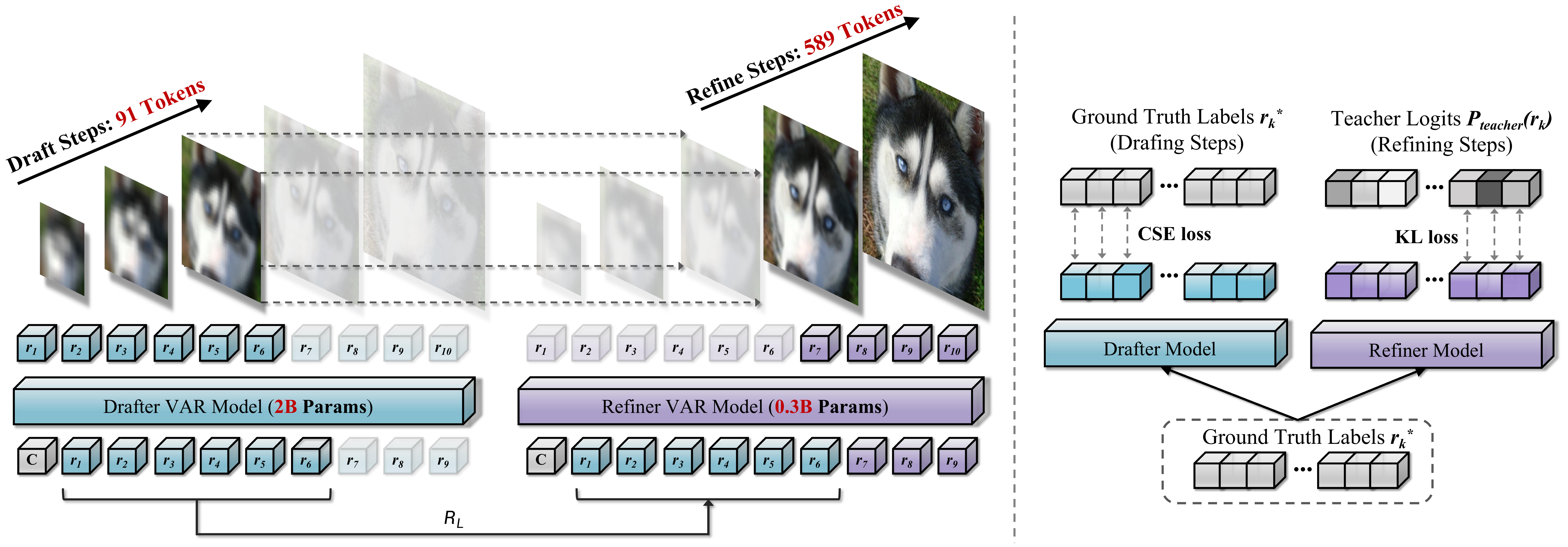}
\caption{Overview of the collaborative decoding process, we use a drafting step $N = 6$ for instance. CoDe uses a large VAR model as the drafter $\epsilon_{\theta_d}$ to generate the token maps $R_L = (r_1, r_2, \ldots, r_N)$ at smaller scales. The small refiner model $\epsilon_{\theta_r}$ then uses $R_L$ as an initial prefix to efficiently predict the remaining token maps $R_H = (r_{N+1}, r_{N+2}, \ldots, r_K)$ at larger scales. Both models are fine-tuned on their designated predictive scales using ground truth labels $r_k^*$ and teacher logits $p_{\text{teacher}}(r_k)$, respectively.}
\label{fig_frame}
\end{figure*}

\subsection{Key Observations}
VAR represents an innovative paradigm specific for visual generation, distinct from traditional autoregressive approaches and introducing many yet unexplored characteristics. In this work, we revisit the entire next-scale prediction process to uncover specific properties that can be optimized to reduce computational redundancy and improve inference efficiency.


\noindent \textbf{Observation 1:} \emph{As the predicted scale becomes larger, the need for parameters reduces significantly.} While VAR models exhibit strong scalability, the effectiveness of increasing parameter counts varies greatly in predicting various scales. In Figure \ref{fig_obs}. (a), we present the differences in class-conditional generation (ImageNet-256 \cite{deng2009imagenet}) quality when using VAR models of different sizes at each scale. At the initial small scales, increasing model parameters leads to noticeable improvements in generation quality. However, as the predicted scale gradually progresses, the impact of additional parameters becomes minimal. At the final scale—which accounts for 38\% tokens of the entire sequence—we observe that the performance of the 2B model is even nearly equivalent to that of the 0.3B model. These findings suggest that as the predicted scale increases, the parameter demand for accurate token predictions declines markedly, revealing substantial computational redundancy in the current VAR inference process at large scales.

\noindent \textbf{Observation 2:} \emph{The generative patterns are exclusive between small and large scales.}
For VAR models, the content generated at small and large scales varies significantly. As demonstrated by the Fourier analysis in Figure \ref{fig_obs} (b), the feature maps produced by the first three scales contain primarily low-frequency components, while the last three scales focus on high-frequency components. To further validate these distinct generative patterns, we conducted a perturbation fine-tuning experiment. Using a pre-trained VAR-d16 model, we applied CSE loss only to tokens in the largest three scales and fine-tuned for just 1\% of the original training epochs. This minor fine-tuning at large scales led to a complete collapse of the model's global modeling capacity at small scales, with the FID increasing from 3.30 to 21.93 and the IS score dropping from 277 to 88. These results demonstrate that VAR models perform entirely distinct generative tasks at small and large scales, with minimal overlap. Training a single VAR model to predict across all scales results in significant mutual interference between learning low-frequency and high-frequency modeling capabilities, making parameter optimization challenging.

In sum, the key insights from our observations can be concluded as follows: (1) At large scales, VAR models require significantly fewer parameters for accurate token prediction, leading to considerable computational redundancy. (2) Training a single model to generate across all scales causes mutual interference between small and large scales, hindering parameter optimization due to conflicting low-frequency and high-frequency learning tendencies.

\subsection{Collaborative Decoding}
Based on the above insights, we propose a simple yet pretty efficient decoding method for next-scale prediction called collaborative decoding. As presented in Figure \ref{fig_frame}, CoDe decomposes the next-scale prediction process into a collaboration between a large drafter model and a small refiner model. The drafter is the original large VAR model with 2B parameters, while the refiner is a lightweight 0.3B VAR model, sharing the same transformer-based architecture but with reduced width and depth. The drafter is responsible for generating coarse, low-frequency global structures at smaller scales to draft the image, while the refiner continually predicts the high-frequency details at larger scales to refine the image.

\noindent \textbf{Model Collaboration.} To generate an image $im$, original VAR model needs to auto-regressively generate \( K \) multi-scale token maps \( R = (r_1, r_2, \ldots, r_K) \) with progressively finer resolutions, where each token map \( r_k \in [V]^{h_k \times w_k} \) represents the token map at scale \( k \). In CoDe, we decompose the auto-regressive steps into $N$ drafting steps and $K-N$ refining steps. In the initial drafting phase where computation is sparse but additional parameters provide significant benefits, we serve a large VAR model as the drafter $\epsilon_{\theta_d}$, generating the initial set of low-frequency token maps $R_L =  (r_1, r_2, \ldots, r_N)$ up to scale \( N \), where \(N < K \). These token maps predict the global structure of the image, serving as a prefix for the remaining finer-scale token maps. The drafting process can be represented as:
\begin{equation}
p_{\theta_d}(r_1, r_2, \ldots, r_N) = \prod_{k=1}^N p_{\theta_d}(r_k \mid r_1, r_2, \ldots, r_{k-1}).
\end{equation}
After drafting, the KV cache of the large model is released, significantly optimizing memory usage.

In the refining stage where the computation is intensive but the parameter demand is low, we serve a small VAR model as the refiner $\epsilon_{\theta_r}$, employing the drafter's predictions $R_L$ as a prefix and focus on refining the details by generating the remaining high-frequency token maps $R_H = (r_{N+1}, r_{N+2}, \ldots, r_K)$ up to scale \( K \), where \( r_K \) matches the resolution of the original feature map. Notably, predicting $r_{N+1}$ requires an attention mask, as the refiner model lacks the KV cache for the first $N$ scales.
The probability distribution of refining steps is formulated as:
\begin{equation} \begin{split}
&p(r_{N+1}, r_{N+2}, \ldots, r_K \mid R_L) \\
= &\prod_{k=N+1}^K p(r_k \mid R_L, r_{N+1}, r_{N+2}, \ldots, r_{k-1}). \end{split}
\end{equation}
Finally, the generated images $im$ are reconstructed from both drafting maps $R_L$ and refining maps $R_H$ as follows:
\begin{equation}
im = D(Q(R_L,R_H),
\end{equation}
where $Q(.)$ is a residual-style quantization function, and $D(.)$ is a multi-scale VQVAE decoder. Figure \ref{fig_obs}. (c) demonstrates the remarkable effectiveness of model collaboration under training-free conditions. With a large model as the drafter, the small refiner only results in a slight and gentle increase in FID. Conversely, with a small model as the drafter, even assigning the last 80\% tokens to a large refiner fails to improve performance. This also verifies Observation 1: small scales require more model capacity, whereas the parameter demands for large scales are low. 

The model collaboration method effectively reduces the computational redundancy of VAR and maintains generation quality comparable to the original model, while significantly offering faster speed and reduced memory usage.

\noindent \textbf{Specialized Fine-Tuning.} Given the exclusive generation patterns between drafting scales and refining scales, we propose the specialized fine-tuning to further specialize drafter and refiner models by fine-tuning them solely on their respective predictive scales, thereby avoiding training interference and enhancing generation quality.

For the drafter model $\epsilon_{\theta_d}$, we fine-tune the drafting steps to improve the generation of the initial low-frequency token maps $R_L$. The drafter model is trained using a CSE loss $\text{CSE}(\cdot, \cdot)$ between its generated token distribution $p_{\theta_d}(r_k)$ and the ground truth labels $r_k^*$, defined as:
\begin{equation}
\mathcal{L}_{\text{drafter}} = \sum_{k=1}^N \text{CSE}(p_{\theta_d}(r_k), r_k^*),
\end{equation}
which encourages the drafter to closely match the target global structure at each drafting step.

For the refiner model $\epsilon_{\theta_r}$, we adopt a \emph{Knowledge Distillation} (KD) \cite{hinton2015distilling} approach to partially transfer the knowledge from a larger pre-trained VAR model. The KD loss function for epoch $ep$ can be expressed as:
\begin{equation}
\small
\mathcal{L}_{\text{refiner}}^{(ep)} = \sum_{k=1}^K \big(\lambda_{ep} \cdot \mathbf{1}_{[k \leq N]} + \mathbf{1}_{[k > N]}\big) \, \text{KL}(p_{\theta_r}(r_k) \parallel p_{\text{teacher}}(r_k)),
\end{equation}
where $p_{\text{teacher}}(\cdot)$ is the distribution predicted by the larger teacher VAR model, $\text{KL}(\cdot, \cdot)$ is the standard KL-divergence loss aligns refiner's output distribution with that of the teacher model, and $\lambda_{ep}$ is a dynamic weighting factor. $\lambda_t$ linearly decreases from one to zero from the initial to the end of finetuning, gradually shifting the KD emphasis from all tokens to the specific refining tokens. This dynamic weighting allows the refiner to specialize in refining the details with a smoother and more stable training process.

The specialized fine-tuning of the drafter and refiner models ensures that each model becomes highly proficient in its respective task. By minimizing interference between small and large scales, this approach allows for more sufficient parameter optimization, resulting in enhanced image generation quality with limited training costs.

\begin{figure*}[ht]
\centering
\includegraphics[width=7.1in]{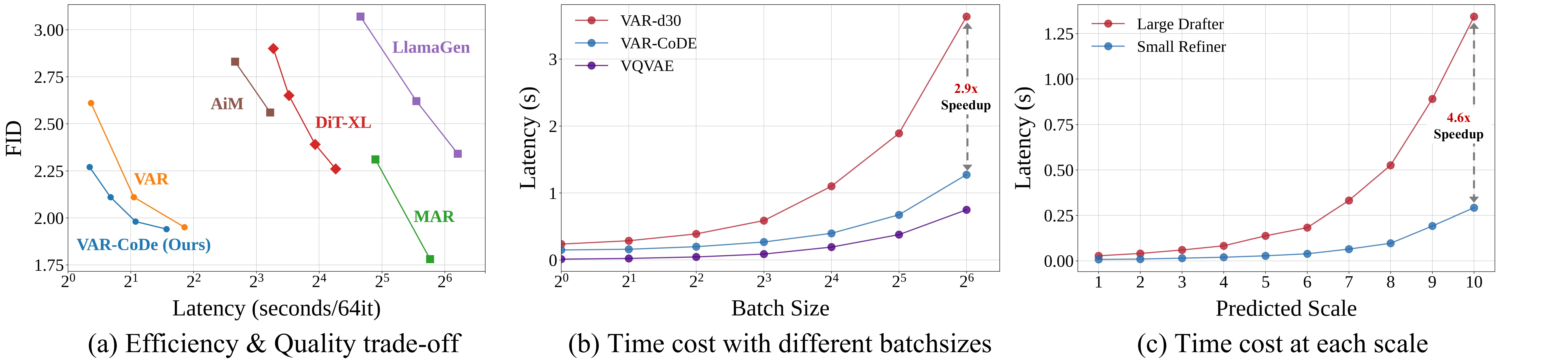}
\caption{(a) Our CoDe demonstrates the optimal efficiency-quality trade-off among all evaluated methods. (b) Inference latency is measured across varying batch sizes for the original VAR-d30, our CoDe (N=6), and the VQVAE decoder. (c) We analyze the time cost associated with parallel decoding at each scale, showing that the refiner model is significantly more efficient than the drafter at larger scales.}
\label{fig_curve_exp}
\end{figure*}

\begin{table*}[!t]
\centering
\caption{Quantitative assessment of the efficiency-quality trade-off across various methods. Inference efficiency is evaluated with a batch size of 64 on NVIDIA L20 GPU, with latency measured excluding VQVAE or VQGAN as it incurs a shared time cost across all methods.}
\small
\begin{tabular}{@{}>{\centering\arraybackslash}m{2.2cm}  | 
*{1}{>{\centering\arraybackslash}m{0.7cm}}
*{1}{>{\centering\arraybackslash}m{1.0cm}}  *{1}{>{\centering\arraybackslash}m{0.9cm}}  *{1}{>{\centering\arraybackslash}m{1.6cm}} *{1}{>{\centering\arraybackslash}m{1.4cm}} *{1}{>{\centering\arraybackslash}m{1.5cm}}| *{2}{>{\centering\arraybackslash}m{0.8cm}} *{2}{>{\centering\arraybackslash}m{1.15cm}} @{}}
\hline
\toprule
\multirow{2}{*}{\bf Method} & \multicolumn{6}{c|}{\bf Inference Efficiency} & \multicolumn{4}{c}{\bf Generation Quality}  \\
\cmidrule(lr){2-7} \cmidrule(lr){8-11}
& {\bf \#Steps} & {\bf Speedup$\uparrow$} & {\bf Latency$\downarrow$}  & {\bf Throughput$\uparrow$}& {\bf \#Param} & {\bf Memory$\downarrow$} & {\bf FID $\downarrow$} & {\bf IS $\uparrow$} & {\bf Precision$\uparrow$} & {\bf Recall$\uparrow$}  \\
\midrule
DiT-XL/2   &50 & 0.2x & 19.20s &  3.33it/s& 675M & 11369MB  & 2.26 & 239 & 0.80 & 0.60  \\
MAR-B   &100 & 0.1x & 29.80s &  2.15it/s & 208M & 8725MB  & 2.31 & 282 & 0.82 & 0.57  \\
AiM-XL  &256 &0.4x & 9.32s &  6.87it/s & 763M & 20983MB   & 2.56 & 257 & 0.82 & 0.57  \\
LlamaGen-XXL &384  &$<$0.1x & 73.97s &  0.87it/s & 1.4B & 42632MB & 2.34 & 254 & 0.80 & 0.59  \\
\midrule
VAR-d30   &10    &1.0x & 3.62s &  17.71it/s & 2.0B & 39228MB  & 1.95 & 301 & 0.81 & 0.59  \\
VAR-d24   &10    &1.7x & 2.07s &  30.92it/s & 1.0B & 25093MB  & 2.11 & 311 & 0.82 & 0.59  \\
VAR-d20   &10    &2.8x & 1.29s &  49.62it/s & 600M & 17814MB  & 2.61 & 301 & 0.83 & 0.56  \\
\midrule
\textbf{VAR-CoDe N=9} &9+1 &1.2x & 2.97s & 21.54it/s & 2.0+0.3B & 28803MB   & 1.94 & 296 & 0.80 & 0.61 \\
\textbf{VAR-CoDe N=8} &8+2 &1.7x & 2.11s & 30.33it/s &2.0+0.3B & 21019MB   & 1.98 & 302 & 0.81 & 0.60 \\
\textbf{VAR-CoDe N=7} &7+3 &2.3x & 1.60s & 40.00it/s &2.0+0.3B & 19943MB  & 2.11 & 303 & 0.82 & 0.59 \\
\textbf{VAR-CoDe N=6} &6+4 &2.9x & 1.27s & 50.39it/s &2.0+0.3B & 19943MB & 2.27 & 297 & 0.82 & 0.58 \\
\bottomrule
\hline
\end{tabular}
\label{tab:quant}
\end{table*}

\begin{table}[!t]
\centering
\small
\caption{Effect of specialized fine-tuning}
\begin{tabular}{@{}>{\raggedright}m{2.6cm}  | *{2}{>{\centering\arraybackslash}m{0.8cm}}  *{2}{>{\centering\arraybackslash}m{0.8cm}}}
\hline
\toprule
\multicolumn{1}{c|}{\multirow{1}{*}{\bf Discription}} 
  & {\bf N=6} & {\bf N=7} & {\bf N=8} & {\bf N=9} \\
\midrule
CoDe Training-free       & 2.42 & 2.26 & 2.10 & 1.99 \\
CoDe Fine-tuning  &  \textbf{2.27} & \textbf{2.11} & \textbf{1.98} & \textbf{1.94} \\
\bottomrule
\hline
\end{tabular}
\label{tab:finetune}
\end{table}

\section{Experimental Results}

\subsection{Experimental Setup.}
\noindent \textbf{Implementation Details.} 
We evaluated our method's effectiveness on the ImageNet1K \cite{deng2009imagenet} class-conditional generation benchmarks. The proposed CoDe framework involves collaboration between a large and a small VAR model. Specifically, we use the pre-trained VAR-d30 model as the drafter and the pre-trained VAR-d16 model as the refiner. Each model undergoes specialized fine-tuning focused exclusively on its respective predictive scales. For the drafter, we fine-tune it for 5\% of its original training epochs with a base learning rate of 1e-6 and a weight decay of 0.08. The refiner model, on the other hand, is fine-tuned for 25\% of its original training epochs with a base learning rate of 1e-5 and no weight decay. Both models are optimized using the AdamW \cite{kingma2014adam} optimizer with a batch size of 1024, achieved through gradient accumulation. The fine-tuning was conducted on 4 NVIDIA L20 GPUs.

\noindent \textbf{Evaluations.} 
We evaluated our method on ImageNet1K-256 generation, focusing on both quality and efficiency metrics. For quality assessment, we used standard metrics such as FID, Inception Score (IS), Precision, and Recall. For other baselines, we use their default sampling methods. For CoDe, we reduced the sampled top-k from 900 to 600 due to improved token prediction accuracy after fine-tuning, while keeping the default top-p at 0.96. To compensate for reduced diversity, we introduced a temperature coefficient of $t=1.1$ for sampling on the smallest 7 scales.
Efficiency was measured through inference latency, throughput, memory consumption, and parameter count to provide a comprehensive comparison. Notably, the speed measurements exclude the VQGAN decoder, which contributes minimally to overall runtime and is a shared component across all methods. All efficiency tests were conducted on a single NVIDIA L20 GPU without additional optimizations such as FlashAttention \cite{dao2022flashattention}, using PyTorch 2.1 \cite{paszke2019pytorch} and FP16.

\subsection{Main Results.}
\noindent \textbf{Quality-Efficiency Trade-off.} 
We evaluated the quality-efficiency trade-off of our proposed method (CoDe) against original VAR models \cite{tian2024visual}, state-of-the-art AR models, and diffusion models. The AR models included GPT-style LlamaGEN \cite{sun2024autoregressive}, Mamba-based Aim \cite{li2024scalable}, and MAE-style MAR \cite{li2024autoregressive}. For diffusion models, we employed the widely used DiT models \cite{peebles2023scalable}. To measure the trade-offs, we used series models with different parameter counts for VAR, LlamaGEN, Aim, and MAR, while for DiT we varied the DDIM \cite{song2020denoising} steps to generate the efficiency-quality curve.

\begin{figure*}[ht!]
\centering
\includegraphics[width=\linewidth]{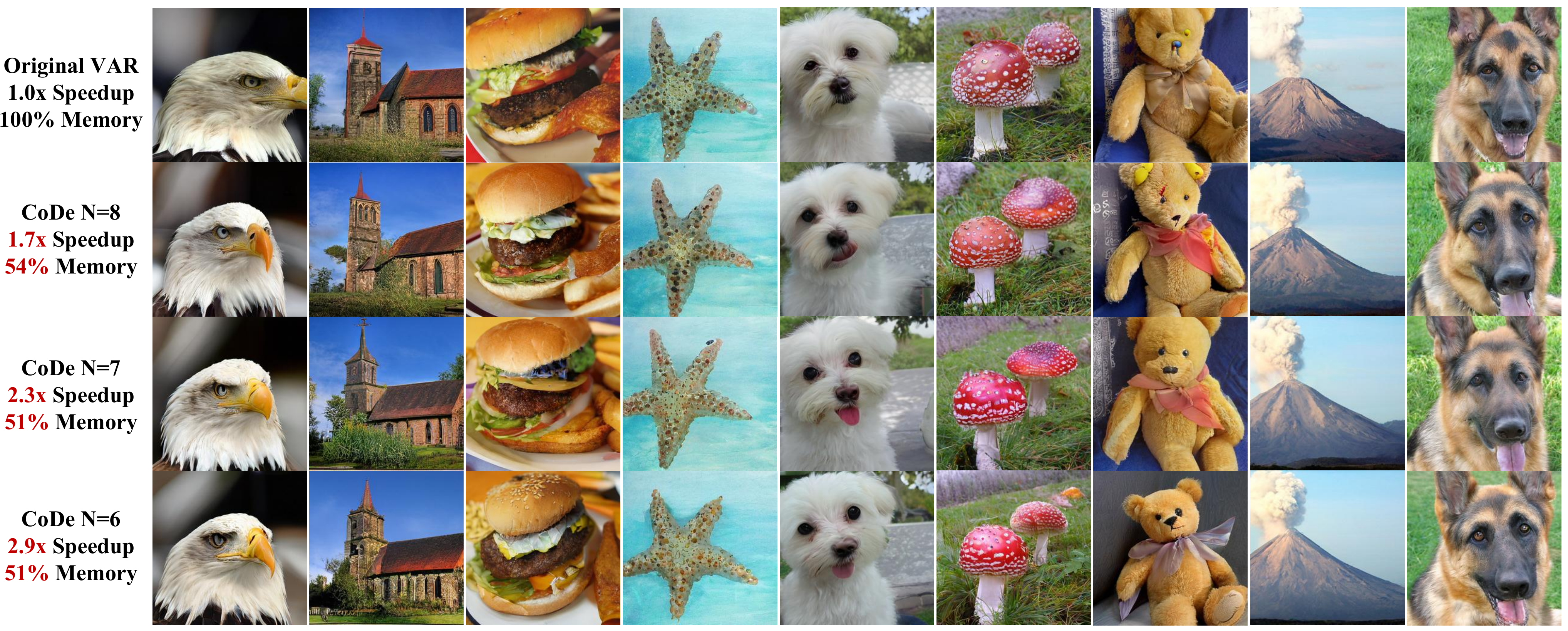}
\caption{Qualitative comparison between the original VAR-d30 model and our proposed CoDe model, with different drafting steps.}
\label{fig_compare}
\end{figure*}

As illustrated in Table \ref{tab:quant}, We analyzed CoDe's performance under different drafting steps N. Compared to traditional AR models, our proposed VAR-CoDe (N=6) achieves 60 times faster inference than LlamaGEN-XXL, while also surpassing its generation quality. Additionally, in comparison to diffusion models, VAR-CoDe (N=6) is 15 times faster than DiT-XL/2, while maintaining the same level of quality. Compared to the original VAR-d30, CoDe (N=8) achieves a 1.7x speedup and reduces memory consumption by 50\%, with only a negligible FID increase from 1.95 to 1.98. When the drafting stage is reduced to just 6 steps, CoDe (N=6) achieves a 2.9x speedup, reaching a throughput of over 50it/s, while maintaining a low FID of 2.27. This is a speed unmatched by any other existing methods. 
Compared to other VAR models with fewer parameters (VAR-d24 and VAR-d20), our method achieves significantly better quality while maintaining the same speedup ratio.

As shown in Figure \ref{fig_curve_exp} (a), VAR-CoDe achieves the best efficiency-quality trade-off compared to all other methods, effectively solving the inefficiencies introduced by the extended sequence length in the original VAR paradigm.

\noindent \textbf{Training-Free vs Specialized Finetuning.}
Our CoDe framework utilizes a large drafter model alongside a smaller refiner model for progressive inference. It can operate in a training-free manner by directly using pre-trained VAR-d30 and VAR-d16 models as the drafter and refiner, respectively. Alternatively, we can perform specialized fine-tuning to further enhance the models' performance at their respective scales, with limited training cost. Table \ref{tab:finetune} compares the results of training-free CoDe with the specialized fine-tuned version. Even without additional training, CoDe demonstrates competitive performance, outperforming VAR-d24 and VAR-d20 models with the same speedup ratio. With specialized fine-tuning, CoDe's performance improves significantly, even achieving a slight quality enhancement over the original model at a 1.2x acceleration ratio. These results demonstrate the superiority of our approach over the conventional VAR paradigm and highlight the gains from specialized fine-tuning for optimal parameter utilization

\noindent \textbf{Qualitative Results.} 
We provide an extensive qualitative comparison between the original VAR-d30 model and our proposed CoDe, with varying drafting steps $N = \{6, 7, 8\}$. As illustrated in Figure \ref{fig_compare}, our approach achieves significant speedup and substantial memory optimization, with only minimal quality degradation that is nearly imperceptible to the human eye. Even at a speedup rate of 2.9 times, the generated images maintain exceptionally high quality and accurate semantic information. It is essential to note that the objective of CoDe is to enhance the efficiency of VAR's inference process while preserving generation quality, rather than mirroring the exact outputs of the original model. Through specialized fine-tuning, CoDe's draft model exhibits higher predictive accuracy compared to the original model. This sometimes results in a different global structure from the original output, yet the image quality remains consistently high or even better.

\begin{figure*}[ht!]
\centering
\includegraphics[width=7in]{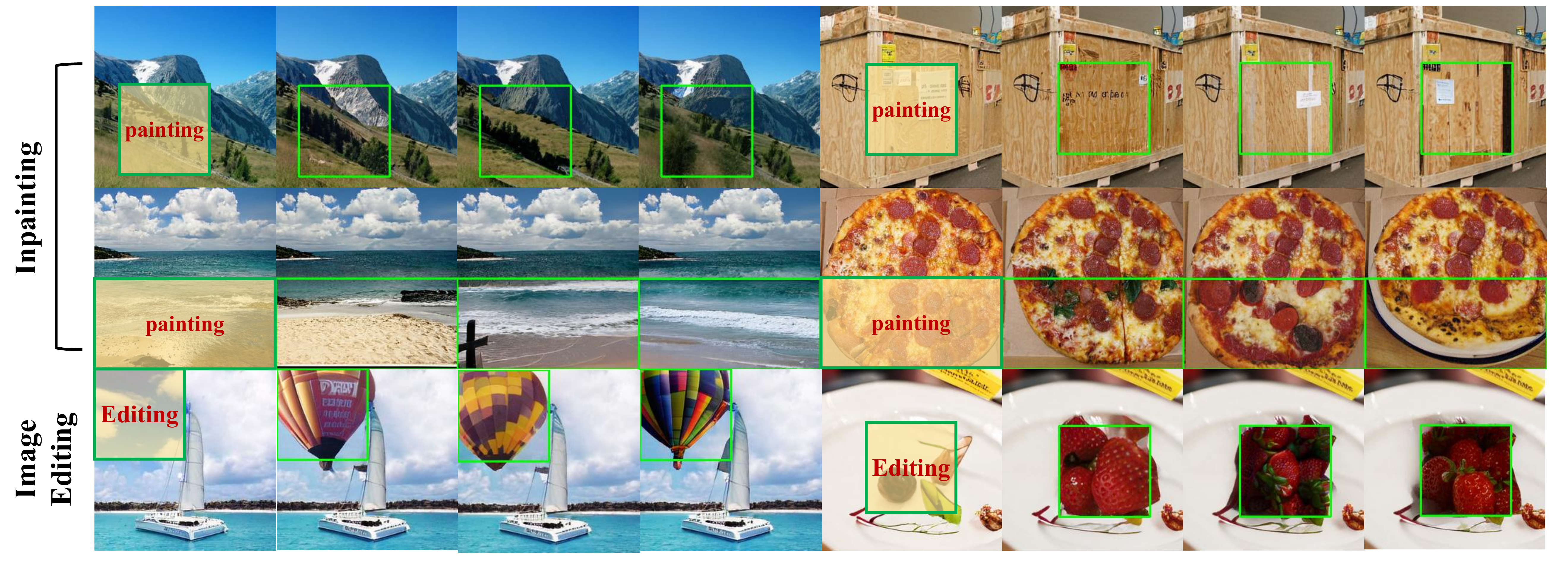}
\caption{Qualitative results of CoDe's zero-shot generalization on image inpainting and image editing.}
\label{fig_zero}
\end{figure*}

\noindent \textbf{Zero-shot Task Generalization.} 
To evaluate the zero-shot generalization capability of CoDe, we conducted additional experiments on zero-shot class-conditional inpainting and image editing. During image inpainting, we applied teacher forcing by providing the ground truth tokens outside the masked area, allowing the model to generate tokens solely within the mask. Notably, class-conditional information was introduced to the model. In the image editing task, CoDe was restricted to generating tokens within a given bounding box based on a specific class label. Figure \ref{fig_zero} illustrates the qualitative results of CoDe's zero-shot performance (N=8). Our approach demonstrates strong zero-shot generalization without any additional training on these downstream tasks.

\subsection{Efficiency Analysis.}
\noindent \textbf{Time Cost Analysis.} 
We first present the time cost (bs=64) for each decoding step of the VAR-d30 model in Figure \ref{fig_curve_exp} (c). Notably, the computational complexity across the ten decoding steps of VAR is highly non-uniform. The last three decoding steps alone account for 64\% of the total inference time, with latency increasing quadratically as token map resolution grows. 
This highlights that addressing the efficiency bottleneck of large-scale predictions is crucial to enhance inference efficiency.
Our solution is to replace the large VAR model with a smaller one for the last few steps, as fewer parameters are needed for large-scale token maps. This small refiner is substantially faster than the original large drafter at these scales, achieving a 4.6x speed improvement in the final and most computationally demanding step. This change dramatically accelerates the entire inference process compared to the original VAR model.

\begin{table}[!t]
\centering
\scriptsize
\caption{Memory usage comparison across different batch sizes}
\begin{tabular}{@{}>{\centering\arraybackslash}m{1.5cm}  | *{2}{>{\centering\arraybackslash}m{1.3cm}}  *{2}{>{\centering\arraybackslash}m{1.25cm}}}
\hline
\toprule
\multirow{2}{*}{\bf Method} & \multicolumn{4}{c}{\bf Memory Consumption$\downarrow$} \\
\cmidrule(lr){2-5}
  & {\bf Running} & {\bf KV Cache} & {\bf Params}&{\bf Total} \\
\midrule
 VAR (bs=8)       & 314MB & 3595MB & \textbf{8089MB} & 12002MB \\
\textbf{+CoDe }  &  \textbf{284MB} & \textbf{1023MB} & 9275MB & \textbf{10619MB} \\
\midrule
 VAR (bs=16)      &  615MB & 7191MB & \textbf{8089MB} & 15901MB \\
\textbf{+CoDe} &  \textbf{557MB} &  \textbf{2056MB} & 9275MB & \textbf{11951MB} \\
\midrule
 VAR (bs=32)    &  1216MB & 14345MB & \textbf{8089MB} & 23662MB \\
\textbf{+CoDe}  & \textbf{1103MB} &  \textbf{4083MB} & 9275MB& \textbf{14614MB} \\
\midrule
 VAR (bs=64)    &  2420MB & 28707MB & \textbf{8089MB} & 39228MB \\
\textbf{+CoDe}  & \textbf{2195MB} &  \textbf{8160MB} & 9275MB & \textbf{19943MB} \\
\midrule
 VAR (bs=128)   & OOM(0.48GB)  & OOM(57GB) & OOM(0.80GB) & OOM(70GB) \\
\textbf{+CoDe}  &  \textbf{4380MB} &  \textbf{16320MB} & 9275MB & \textbf{30598MB} \\
\bottomrule
\hline
\end{tabular}
\label{tab:memory}
\end{table}

Next, we present the speedup effect of CoDe (N=6) across different batch sizes. As shown in Figure \ref{fig_curve_exp}.(b), CoDe achieves a speedup of 1.6x for batch size 1 but reaches up to 2.9x for batch size 64. The reduced speedup at smaller batch sizes is due to low GPU utilization when generating 256x256 images, which limits the acceleration potential. However, as GPU utilization increases with larger batch sizes, the speedup of CoDe also improves significantly. This indicates that CoDe is well-suited for computationally intensive image generation tasks, and holds great potential for enhancing efficiency in high-resolution image generation under the VAR paradigm. Additionally, we report the time cost of the VQVAE decoder in the figure, which is a constant shared cost across all methods.

\noindent \textbf{Memory Consumption Analysis.} 
In Table \ref{tab:memory}, we provide a detailed analysis of memory usage during the VAR inference process. The KV cache of the VAR model is the largest memory consumer, requiring 12 times the memory needed for the model's decoding operation due to the significantly extended sequence length in the VAR paradigm. Our proposed CoDe effectively addresses this KV cache memory challenge. Specifically, we use the large VAR model only for predicting the first 91 tokens, which represent just 13\% of the total sequence. After this, the KV cache in the drafter is released, and the remaining computation is handled by a smaller model. This approach drastically reduces the KV cache memory requirements, compressing it to approximately 28\% of the original VAR-d30. Although the refiner model adds a small number of additional parameters, this overhead is minimal compared to the major optimization achieved in KV cache storage. As a result, the total memory usage of CoDe is significantly lower than the original model.
Moreover, as batch size increases, the memory savings with CoDe become even more pronounced, underscoring its potential to support efficient high-resolution image generation within the VAR paradigm.

\section{Conclusion}
\label{sec:conclusion}
This work presents CoDe, a novel method designed for efficient decoding in visual auto-regressive modeling. CoDe effectively mitigates the significant memory overhead and computational redundancy typically associated with the prolonged sequences of next-scale predictions. Through extensive experimentation, our method demonstrates a superior efficiency-quality trade-off, establishing a new benchmark for efficient, high-quality image generation. 

{
    \small
    \bibliographystyle{ieeenat_fullname}
    \bibliography{main}
}

\clearpage
\appendix
\setcounter{page}{1}
\maketitlesupplementary

\noindent In this document, we provide supplementary materials that extend beyond the scope of the main manuscript, constrained by space limitations. These additional materials include:
\begin{itemize}
\item We provide more quantitative analysis results to further illustrate our approach;
\item We offer more qualitative comparisons for visualization; 
\item We discuss the limitations of our approach and look into future work.
\end{itemize}

\section{Additional Quantitative Results}
In this section, we present additional quantitative analyses to further substantiate our approach.

\noindent \textbf{Impact of Increasing Model Parameters.}  
To validate \textbf{Observation 1}, we analyze the effect of varying model sizes on class-conditional image generation quality using ImageNet-256 \cite{deng2009imagenet}. Specifically, we evaluate the impact of model size at the $k$-th scale by predicting the token map $r_k$ with four Visual Autoregressive (VAR) models \cite{tian2024visual} of different parameter sizes (2B, 1B, 0.6B, and 0.3B). For all other scales ($r_1, r_2, \dots, r_{k-1}, r_{k+1}, \dots, r_{10}$), the largest VAR-d30 model is used for generation. Detailed quantitative results are summarized in Table \ref{tab:scaling}. Our results reveal that increasing model parameters at the earlier scales yields significant improvements in generation quality. However, as the scales progress, the marginal benefits of larger models diminish. At the final scale—responsible for 38\% of the sequence tokens—we observe that the performance of the 2B model is nearly identical to that of the 0.3B model. This indicates that as the predicted scale increases, the demand for model parameters to ensure accurate token predictions decreases substantially. These findings highlight significant computational redundancy in the current VAR inference process at larger scales.

\begin{table}[!t]
\centering
\small
\renewcommand{\arraystretch}{1.1}
\caption{Impact of increasing parameters across scales}

\begin{tabular}{@{}>{\centering\arraybackslash}m{0.7cm}  | *{1}{>{\centering\arraybackslash}m{1.0cm}} |  *{2}{>{\centering\arraybackslash}m{0.9cm}}  *{2}{>{\centering\arraybackslash}m{1.2cm}}}

\hline
\toprule
{\bf Scale} &{\bf Params} & {\bf FID $\downarrow$} & {\bf IS $\uparrow$} & {\bf Precision$\uparrow$} & {\bf Recall$\uparrow$} \\
\midrule
2&  0.3B& 2.23 & 291  &0.8122  &0.5895  \\
2&  0.6B& 2.13 & 292 & 0.8078 &0.5947  \\
2&  1.0B& 2.04 & 295 & 0.8107 &0.6027  \\
2&  2.0B&  1.95  &301  &0.8107  &0.5945  \\
\midrule
3&  0.3B& 2.35 &283  &0.8064  &0.5864  \\
3&  0.6B& 2.21 &290  &0.8047  &0.5967  \\
3&  1.0B& 2.09 &295  &0.8074  &0.5940  \\
3&  2.0B&  1.95  &301  &0.8107  &0.5945  \\
\midrule
4&  0.3B& 2.27  &290  &0.8086  & 0.5953 \\
4&  0.6B& 2.18 &293  &0.8068  & 0.5924 \\
4&  1.0B& 2.13 &296  &0.8061  & 0.5983 \\
4&  2.0B& 1.95  &301  &0.8107  &0.5945  \\
\midrule
5&  0.3B&2.17  &296  &0.8119  &0.5936  \\
5&  0.6B&2.13  &298  &0.8087  &0.5948  \\
5&  1.0B&2.10  &301  &0.8087  &0.6025  \\
5&  2.0B&1.95  &301  &0.8107  &0.5945  \\
\midrule
6&  0.3B&2.09  &301  &0.8119  &0.5984  \\
6&  0.6B&2.05 &304  &0.8100  &0.5976  \\
6&  1.0B&2.05  &305  &0.8089  &0.5999  \\
6&  2.0B&1.95  &301  &0.8107  &0.5945  \\
\midrule
7&  0.3B&2.09  &302  &0.8067  &0.6010  \\
7&  0.6B&2.05  &305  &0.5095  &0.6061  \\
7&  1.0B&2.04  &307  &0.8077  &0.6008  \\
7&  2.0B&1.95  &301  &0.8107  &0.5945  \\
\midrule
8&  0.3B&2.08  &304  &0.8135  &0.5978  \\
8&  0.6B&2.04  &308  &0.8110  &0.6024  \\
8&  1.0B&2.02  &307  &0.8094  &0.6038  \\
8&  2.0B&1.95  &301  &0.8107  &0.5945  \\
\midrule
9&  0.3B&2.02  &304  &0.8133  &0.6059  \\
9&  0.6B&2.01  &307  &0.8121  &0.5948  \\
9&  1.0B&2.00  &307  &0.8097  &0.6011  \\
9&  2.0B&1.95  &301  &0.8107  &0.5945  \\
\midrule
10&  0.3B&1.99  &306  &0.8120  &0.5978  \\
10&  0.6B&1.97  &305  &0.8102  &0.6053  \\
10&  1.0B&1.98  &303  &0.8102  &0.6053  \\
10&  2.0B&1.95  &301  &0.8107  &0.5945  \\
\bottomrule
\hline
\end{tabular}
\label{tab:scaling}
\end{table}

\noindent \textbf{Training-Free Performance of CoDe.}  
The proposed CoDe framework employs a large drafter model in conjunction with a smaller refiner model for progressive inference. Notably, it can operate in a training-free manner by leveraging pre-trained VAR-d30 and VAR-d16 models as the drafter and refiner, respectively. Table \ref{tab:free} presents the performance of training-free CoDe across various drafting step settings $N = \{1, 2, 3, 4, 5, 6, 7, 8, 9\}$. Even without additional training, CoDe achieves competitive performance, surpassing the VAR-d24 and VAR-d20 models while maintaining the same speedup ratio.

\begin{figure*}[h!]
\centering
\includegraphics[width=\linewidth]{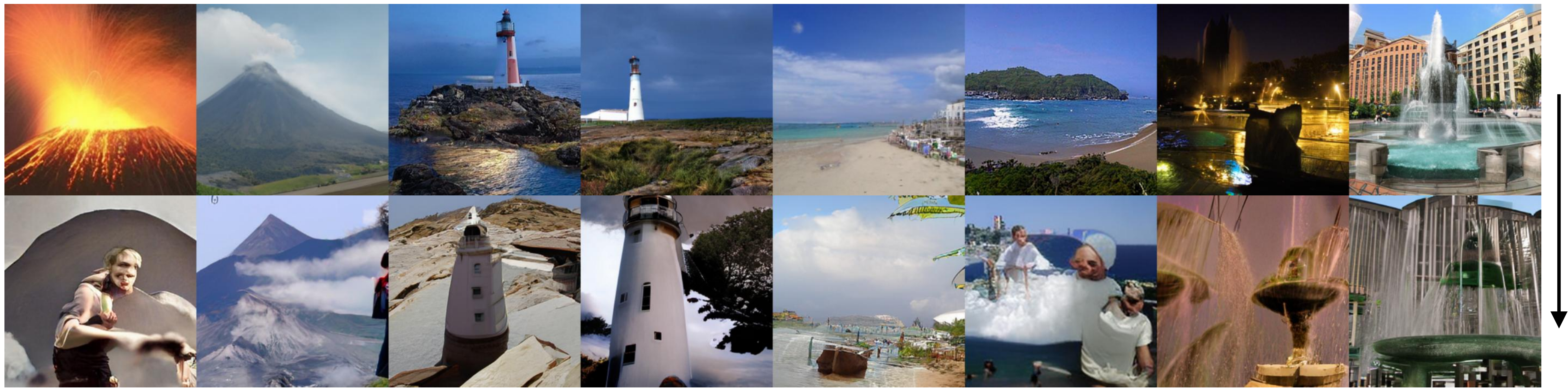}
\caption{Up: images generated by the original VAR-d16 models. Down: images generated by the perturbation fine-tuned VAR-d16.}
\label{fig_perturb}
\end{figure*}

\begin{table*}[!h]
\centering
\caption{No reference metrics for additional image quality assessments.}
\small
\renewcommand{\arraystretch}{1.1} 
\begin{tabular}{@{}>{\centering\arraybackslash}m{2.2cm}  | 
*{1}{>{\centering\arraybackslash}m{0.7cm}}
*{1}{>{\centering\arraybackslash}m{1.0cm}}  *{1}{>{\centering\arraybackslash}m{0.9cm}}  *{1}{>{\centering\arraybackslash}m{1.6cm}} *{1}{>{\centering\arraybackslash}m{1.4cm}} *{1}{>{\centering\arraybackslash}m{1.5cm}}| *{1}{>{\centering\arraybackslash}m{1.4cm}} *{1}{>{\centering\arraybackslash}m{1.6cm}} *{1}{>{\centering\arraybackslash}m{1.0cm}} @{}}
\hline
\toprule
\multirow{2}{*}{\bf Method} & \multicolumn{6}{c|}{\bf Inference Efficiency} & \multicolumn{3}{c}{\bf Image Quality Assessment}  \\
\cmidrule(lr){2-7} \cmidrule(lr){8-10}
& {\bf \#Steps} & {\bf Speedup$\uparrow$} & {\bf Latency$\downarrow$}  & {\bf Throughput$\uparrow$}& {\bf \#Param} & {\bf Memory$\downarrow$} & {\bf MUSIQ $\uparrow$} & {\bf CLIPIQA $\uparrow$} & {\bf NIQE$\downarrow$}   \\
\midrule
VAR-d30   &10    &1.0x & 3.62s &  17.71it/s & 2.0B & 40414MB  & 60.72 & 0.6813 & 6.1739  \\
\midrule
\textbf{VAR-CoDe N=9} &9+1 &1.2x & 2.97s & 21.54it/s & 2.0+0.3B & 28803MB   & 60.78 & 0.6818 & 6.1024  \\
\textbf{VAR-CoDe N=8} &8+2 &1.7x & 2.11s & 30.33it/s &2.0+0.3B & 21019MB   & 60.79 & 0.6812 & 6.0849 \\
\textbf{VAR-CoDe N=7} &7+3 &2.3x & 1.60s & 40.00it/s &2.0+0.3B & 19943MB  & 60.82 & 0.6800 & 6.1247  \\
\textbf{VAR-CoDe N=6} &6+4 &2.9x & 1.27s & 50.39it/s &2.0+0.3B & 19943MB & 60.76 & 0.6808 & 6.1490 \\
\bottomrule
\hline
\end{tabular}
\label{tab:iqa}
\end{table*}

\begin{table}[!h]
\centering
\small
\renewcommand{\arraystretch}{1.1}
\caption{The training-free performance of CoDe}

\begin{tabular}{@{}>{\centering\arraybackslash}m{1.9cm}  |  *{2}{>{\centering\arraybackslash}m{0.9cm}}  *{2}{>{\centering\arraybackslash}m{1.2cm}}}

\hline
\toprule
{\bf Configuration}  & {\bf FID $\downarrow$} & {\bf IS $\uparrow$} & {\bf Precision$\uparrow$} & {\bf Recall$\uparrow$} \\
\midrule
\textbf{CoDe N=9}&   1.99 & 306  &0.8120  &0.5978  \\
\textbf{CoDe N=8}&   2.10 & 308  &0.8155  &0.5915  \\
\textbf{CoDe N=7}&   2.25 & 309 & 0.8204 &0.5781  \\
\textbf{CoDe N=6}&   2.42 & 306 & 0.8283 &0.5721  \\
\textbf{CoDe N=5}&   2.56 &303  &0.8313  &0.5660  \\
\textbf{CoDe N=4}&   2.75 &295  &0.8342  &0.5427  \\
\textbf{CoDe N=3}&   2.99 &288  &0.8410  &0.5327  \\
\textbf{CoDe N=2}&   3.19 &283  &0.8433  &0.5179  \\
\textbf{CoDe N=1}&   3.39 &268  &0.8132  &0.5382  \\
\bottomrule
\hline
\end{tabular}
\label{tab:free}
\end{table}

\noindent \textbf{Image Quality Assessment.}
In our paper, we use standard metrics such as FID \cite{heusel2017gans}, Inception Score (IS), Precision, and Recall to evaluate the generation quality. In order to more comprehensively evaluate the quality of generated images, we introduced three image quality assessment (IQA) metrics, including MUSIQ \cite{ke2021musiq}, CLIPIQA \cite{wang2022exploring}, and NIQE \cite{mittal2012making}. MUSIQ, CLIPIQA, and NIQE are three distinct IQA metrics, each with unique approaches and strengths. MUSIQ (Multi-Scale Image Quality) leverages a vision transformer (ViT) \cite{han2022survey} and a multi-scale representation to evaluate global aesthetics and local distortions, making it effective for diverse image types, including high-resolution and non-standard aspect ratios. CLIPIQA utilizes the pre-trained CLIP \cite{radford2021learning} model, which combines semantic understanding from large-scale image-text training to assess image quality in a context-aware manner, excelling in tasks aligned with human perception. In contrast, NIQE (Natural Image Quality Evaluator) is a no-reference metric that models natural scene statistics (NSS) using a multivariate Gaussian distribution to measure deviations from high-quality natural image properties. While MUSIQ and CLIPIQA excel in leveraging learned features for state-of-the-art performance, NIQE stands out for its simplicity, computational efficiency, and independence from reference images, though it may struggle with unnatural or heavily edited content. Together, these metrics cater to diverse IQA needs, from deep-learning-based evaluations to lightweight statistical assessments. As shown in Table \ref{tab:iqa}, our CoDe method achieves comparable or even superior generation quality compared to the original VAR-d30. This result further demonstrates the effectiveness of our approach.

\section{More Qualitative Results.}
\begin{figure*}[h]
\centering
\includegraphics[width=\linewidth]{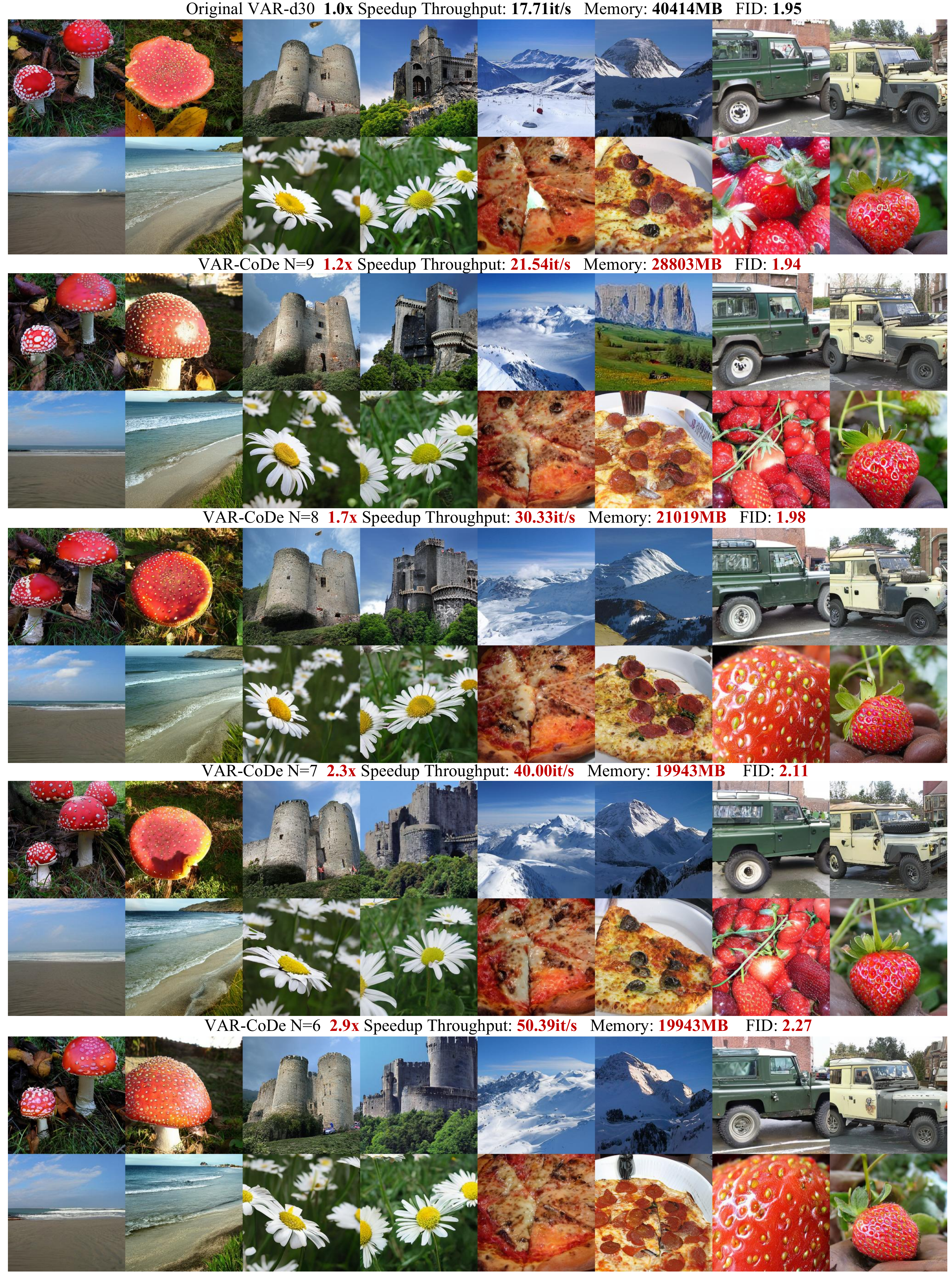}
\caption{Qualitative comparison between the original VAR-d30 model and our proposed CoDe model, with different drafting steps.}
\label{fig_compare2}
\end{figure*}

\begin{figure*}[h]
\centering
\includegraphics[width=\linewidth]{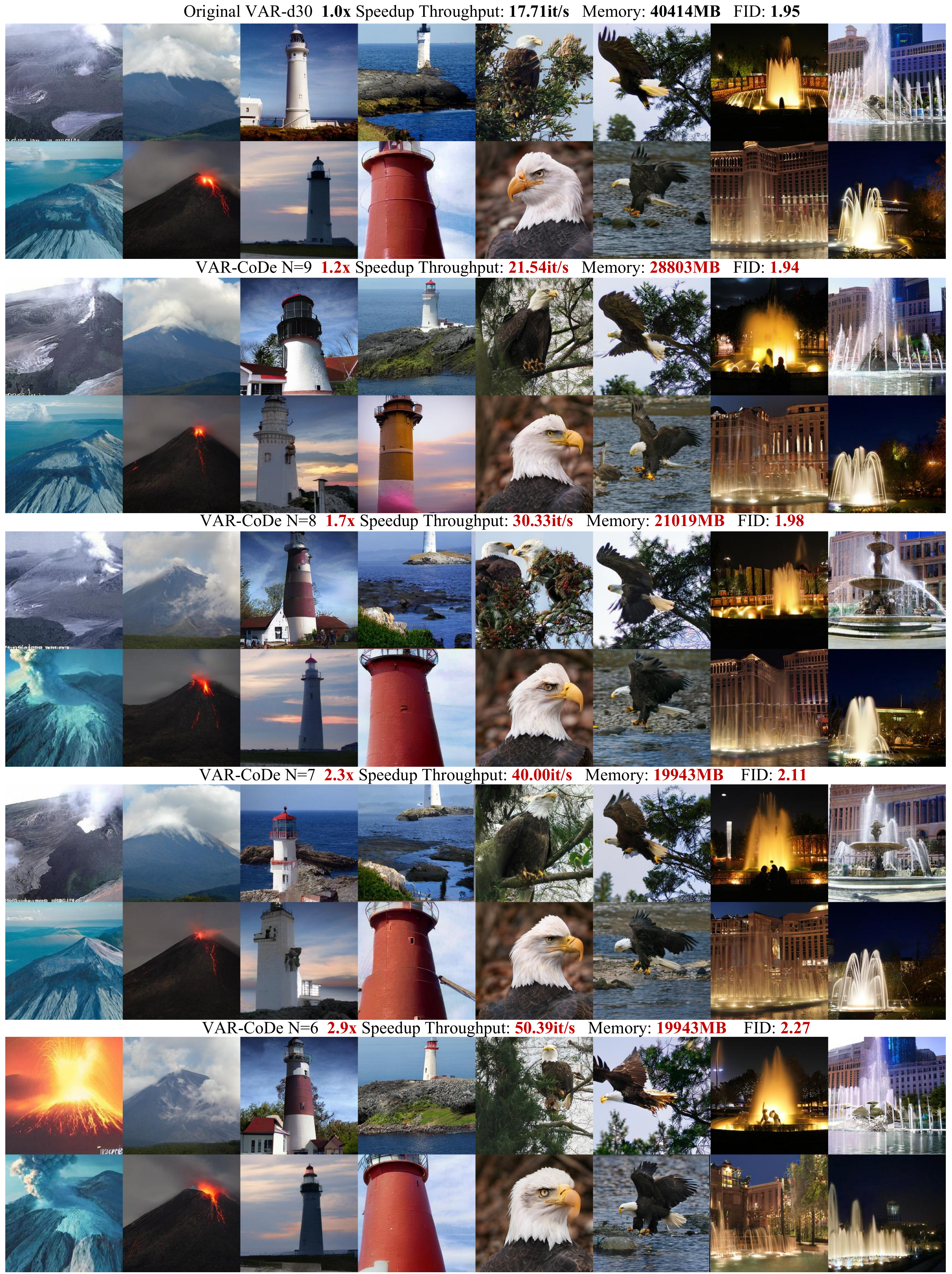}
\caption{Qualitative comparison between the original VAR-d30 model and our proposed CoDe model, with different drafting steps.}
\label{fig_compare1}
\end{figure*}

\noindent \textbf{Additional Qualitative Comparisons.}  
We provide additional qualitative comparisons between the original VAR-d30 model and our proposed CoDe framework, evaluated with varying drafting steps $N = \{6, 7, 8, 9\}$. As shown in Figures \ref{fig_compare2} and \ref{fig_compare1}, CoDe achieves significant speedup and substantial memory optimization, with only minimal quality degradation that is nearly imperceptible to the human eye. Even at a speedup rate of 2.9×, the generated images maintain exceptionally high quality and preserve accurate semantic information. It is important to emphasize that the primary goal of CoDe is to enhance the efficiency of the VAR inference process while maintaining high generation quality, rather than reproducing the exact outputs of the original model. Through specialized fine-tuning, CoDe's drafter model demonstrates superior predictive accuracy compared to the original model, sometimes resulting in a different global structure. Nevertheless, the image quality remains consistently high and, in some cases, even improves over the original outputs.

\noindent \textbf{Qualitative Results of Perturbation Fine-Tuning.}  
In our study, we conducted a perturbation fine-tuning experiment to examine the distinct generative roles of small and large scales. Using a pre-trained VAR-d16 model, we applied the CSE loss exclusively to tokens in the largest three scales and fine-tuned the model for just 1\% of the original training epochs. This minimal fine-tuning at large scales caused a complete collapse of the model's global modeling capacity at small scales, with the FID increasing from 3.30 to 21.93 and the IS score dropping from 277 to 88.
Figure \ref{fig_perturb} illustrates the qualitative results of perturbation fine-tuning. After slight fine-tuning, the VAR-d16 model nearly loses its ability to model global structures. These findings underscore that VAR models undertake entirely distinct generative tasks at small and large scales, with minimal overlap in functionality.

\section{Limitations and Future Work}

\noindent \textbf{Limitations.}  
The core concept of CoDe involves decomposing the next-scale prediction process into a collaboration between a large model and a small model. This approach necessitates the availability of two models with different sizes. If only a single large VAR model is available and faster inference is desired, it becomes necessary to retrain a smaller refiner model. However, since the refiner model can be extremely compact, techniques such as model pruning and knowledge distillation can be applied to limit the additional training cost.

\noindent \textbf{Future Work.}  
This study demonstrates that CoDe significantly reduces inference latency and memory consumption for VAR models. Furthermore, the efficiency gains from CoDe become even more pronounced in computationally intensive scenarios. As a result, CoDe is particularly well-suited for high-resolution image generation tasks based on next-scale prediction. In future work, we aim to explore the application of CoDe in building an efficient VAR model specifically optimized for high-resolution image generation.

\end{document}